\documentclass[conference]{IEEEtran}
\IEEEoverridecommandlockouts
\usepackage{cite}
\usepackage{amsmath,amssymb,amsfonts}
\usepackage{algorithmic}
\usepackage{graphicx}
\usepackage{textcomp}
\usepackage{xcolor}
\usepackage{microtype}
\usepackage{float}
\usepackage{booktabs}
\usepackage{multirow}
\usepackage{adjustbox}
\begin{document}

\title{Subject-independent Classification of Meditative State from the Resting State using EEG}

\author{
\IEEEauthorblockN{Jerrin Thomas Panachakel}\thanks{Personal use of this material is permitted.  Permission from IEEE must be obtained for all other uses, in any current or future media, including reprinting/republishing this material for advertising or promotional purposes, creating new collective works, for resale or redistribution to servers or lists, or reuse of any copyrighted component of this work in other works. Published article \cite{panachakel2024subject}}
\IEEEauthorblockA{\textit{Dept. of Elect. and Comm. Engg.} \\
\textit{College of Engineering Trivandrum}\\
Trivandrum, India \\
jerrin.panachakel@cet.ac.in}
\and
\IEEEauthorblockN{Pradeep Kumar G.}
\IEEEauthorblockA{\textit{Centre for Brain Research} \\
\textit{Indian Institute of Science, Bangalore}\\
Bangalore, India\\
pradeepkg@alum.iisc.ac.in}
\and
\IEEEauthorblockN{Suryaa Seran}
\IEEEauthorblockA{\textit{Language Technologies Institute} \\
\textit{Carnegie Mellon University}\\
Pennsylvania, USA\\
surya.seran@gmail.com}
\and
\hspace{3cm}
\IEEEauthorblockN{Kanishka Sharma}
\IEEEauthorblockA{\hspace{3cm}\textit{School of Psychology and Education} \\
\textit{\hspace{3cm}Rishihood University, Sonipat}\\
\hspace{3cm}Sonipat, India\\
\hspace{3cm}sharmakan17@hotmail.com}
\and
\IEEEauthorblockN{Ramakrishnan Angarai Ganesan}
\IEEEauthorblockA{\textit{Dept. of Heritage Science and Technology} \\
\textit{Indian Institute of Technology, Hyderabad}\\
Hyderabad, India\\
agr@alum.iisc.ac.in}
}
\IEEEoverridecommandlockouts
\IEEEpubid{\makebox[\columnwidth]{979-8-3503-9128-2/24/\$31.00 ~\copyright2024 IEEE\hfill }
\hspace{\columnsep}\makebox[\columnwidth]{ }}
\maketitle

\begin{abstract}
While it is beneficial to objectively determine whether a subject is meditating, most research in the literature reports good results only in a subject-dependent manner. This study aims to distinguish the modified state of consciousness experienced during Rajyoga meditation from the resting state of the brain in a subject-independent manner using EEG data. Three architectures have been proposed and evaluated: The CSP-LDA Architecture utilizes common spatial pattern (CSP) for feature extraction and linear discriminant analysis (LDA) for classification. The CSP-LDA-LSTM Architecture employs CSP for feature extraction, LDA for dimensionality reduction, and long short-term memory (LSTM) networks for classification, modeling the binary classification problem as a sequence learning problem. The SVD-NN Architecture uses singular value decomposition (SVD) to select the most relevant components of the EEG signals and a shallow neural network (NN) for classification. The CSP-LDA-LSTM architecture gives the best performance with 98.2\% accuracy for intra-subject classification. The SVD-NN architecture provides significant performance with 96.4\% accuracy for inter-subject classification. This is comparable to the best-reported accuracies in the literature for intra-subject classification. Both architectures are capable of capturing subject-invariant EEG features for effectively classifying the meditative state from the resting state. The high intra-subject and inter-subject classification accuracies indicate these systems' robustness and their ability to generalize across different subjects.
\end{abstract}


\section{Introduction}

Meditation may be defined as a modified conscious state comprising cognitive components of alertness and attention, combined with a deep state of relaxation. In research experiments studying the effects of meditation, there is a need to objectively determine whether a subject is meditating and, preferably, to quantify the depth of meditation. However, unlike experiments involving physical or intellectual tasks, it is not straightforward to ascertain whether the subject is actually meditating.

This study explores the feasibility of using state-of-the-art machine learning algorithms to classify EEG segments from meditative and resting states. While previous works \cite{tee2020classification,ahani2014quantitative,lin2017using} primarily focus on intra-subject classification (training and testing data from the same participant), this work emphasizes inter-subject classification, where test data come from individuals not in the training set. The inter-subject variability in EEG \cite{saha2020intra} makes this task challenging. A system that successfully classifies meditative and resting states across subjects can also measure meditation depth using classification confidence. High-confidence classification of unseen meditation EEG epochs suggests greater separability from resting states, potentially indicating a deeper meditation state. Systems trained on expert meditators' EEG data can test novice meditators, with high-confidence classification suggesting effective learning of meditation.

The meditation session experience is subjective, and we propose to study the same with architectures capable of learning the common signatures in EEG during meditation across meditators. We propose three architectures for intra-subject and inter-subject classification of meditative-state EEG from the resting-state EEG:
\begin{enumerate}
    \item CSP-LDA architecture where common spatial patterns (CSP) with linear discriminant analysis (LDA) is used.
    \item CSP-LDA-LSTM architecture where CSP with LDA and long short-term memory (LSTM) is used.
    \item SVD-NN architecture where singular value decomposition (SVD) with shallow neural network (NN) is used.
\end{enumerate}

The contributions of this work are as follows.

\begin{enumerate}
 \item This is the maiden work on the inter-subject classification of meditative from resting-state EEG using SVD.
 \item This is the first work on classifying EEG recorded during Rajyoga meditation from that of resting state. Developing such a system is challenging since the subjects meditate with their eyes open throughout the session and hence the alpha band cannot support discrimination.
 \item We have developed architectures that have inter-subject classification performance comparable to the best intra-subject performance reported in the literature.
\end{enumerate}

Studies reported in the literature are limited to intra-subject classification of meditative states. However, any system for classifying non-ordinary states of consciousness including that of meditation \cite{michaels1976evaluation,shapiro1978meditation} requires models with good subject-independent (inter-subject) performance on test data classification. The primary reason for the focus on intra-subject classification in the literature is due to the inferior inter-subject classification results and high inter-subject variability in EEG \cite{saha2020intra}. Our work overcomes this limitation by proposing two architectures that perform comparably to the intra-subject setting. A system with a high inter-subject classification performance may be explored for use cases such as grading the depth of meditation of a novice meditator. Here, the model needs to be trained on expert meditators' EEG data and tested on novice meditators' EEG data. The system can also be used for classifying other modified states of consciousness such as hypnosis and trance.  


\section{Description of the Dataset}

\subsection{Details of the Meditative Subjects}
EEG was recorded from 54 Rajyoga meditators with a mean age of 42\(\pm\)10.1 years and meditation experience from 4 to 43 years (mean = 18 years). None of the subjects had any neurological disorders or had taken alcohol or smoked cigarettes during the six months before the data recording. The protocol was designed as per the tenets of the Declaration of Helsinki and approved by the Indian Institute of Science Human Ethics Committee (IHEC No: 23/24072019). All subjects had provided informed consent.

\subsection{Protocol Used for EEG Data Recording}
\begin{figure}
    \centering
    \includegraphics[width=8cm]{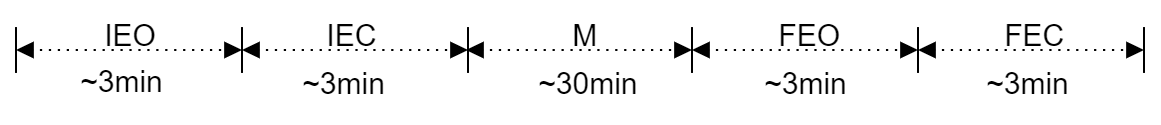}
    \caption{Our research protocol consists of initial eyes open (IEO) and eyes closed (IEC) baselines for 3 minutes each, after which the subjects do seed stage meditation (M) for 30 minutes. The baseline recorded at the end of the experiment also has eyes closed (FEC: final eyes closed) and open (FEO) recordings for 3 minutes each.}
    \label{fig:1}
\end{figure}

EEG was recorded by 10-10 electrode system employing ANT Neuro amplifier, with Waveguard cap of 64 channels. The sampling rate was 500 Hz and the reference electrode was CPz. Electrode impedances were kept under 10 $k\Omega$. More details about the EEG acquisition are available in our previous publications \cite{med-covariance,med-entropy,kumar2021gamma}. Recording sessions included baseline and meditative segments as shown in Fig. \ref{fig:1}. The baseline consists of eyes-open (IEO) and eyes-closed (IEC) recordings, each of duration three minutes. The Rajyoga seed-stage meditation session was for 30 minutes. The current study attempts to classify the eyes-open baseline from the eyes-open Rajyoga meditation.

\subsection{Preprocessing of the EEG}
Spline interpolation was used to downsample the data to 128 Hz, 
and a notch filter, to remove the power line interference. Independent component analysis was employed to eliminate the ocular artifacts. Using EEGLAB \cite{delorme2004eeglab}, the preprocessed EEG data was band-pass filtered into the bands mentioned in Table \ref{table:1}. To balance the data, the initial and final eyes-open baseline recordings and six minutes of meditation data were only included in the analysis.

\begin{table}
    \centering
    \caption{The four EEG frequency bands considered in this study.}
    \label{table:1}
    \begin{tabular}{@{}lcccc@{}}
        \toprule
        \textbf{EEG band} & {$\alpha$} & {$\beta$} & {Low $\gamma$} & {High $\gamma$} \\
        \midrule
        Freq. range (Hz) & 8 - 13 & 13 - 25 & 25 - 45 & 45 - 64 \\
        \bottomrule
    \end{tabular}
    
\end{table}

\section{Description of Architectures}
\subsection{CSP-LDA Architecture}
In this architecture, features are extracted making use of CSP employing Tikhonov regularization (TR-CSP) \cite{Lotte2010}. CSP has previously been utilised for decoding imagined words \cite{dasalla2009single,panachakel2020novel} and motor imagery \cite{sharon2019level}.

The EEG is projected onto the selected spatial vectors and the logarithm of the variance of the resultant signal is used as a feature. If $N_b$ filter pairs are chosen, the feature vector has a dimension of $2N_b$. LDA is used as the classifier.
\begin{figure}[h!]
    \centering
    \includegraphics[width = 8cm]{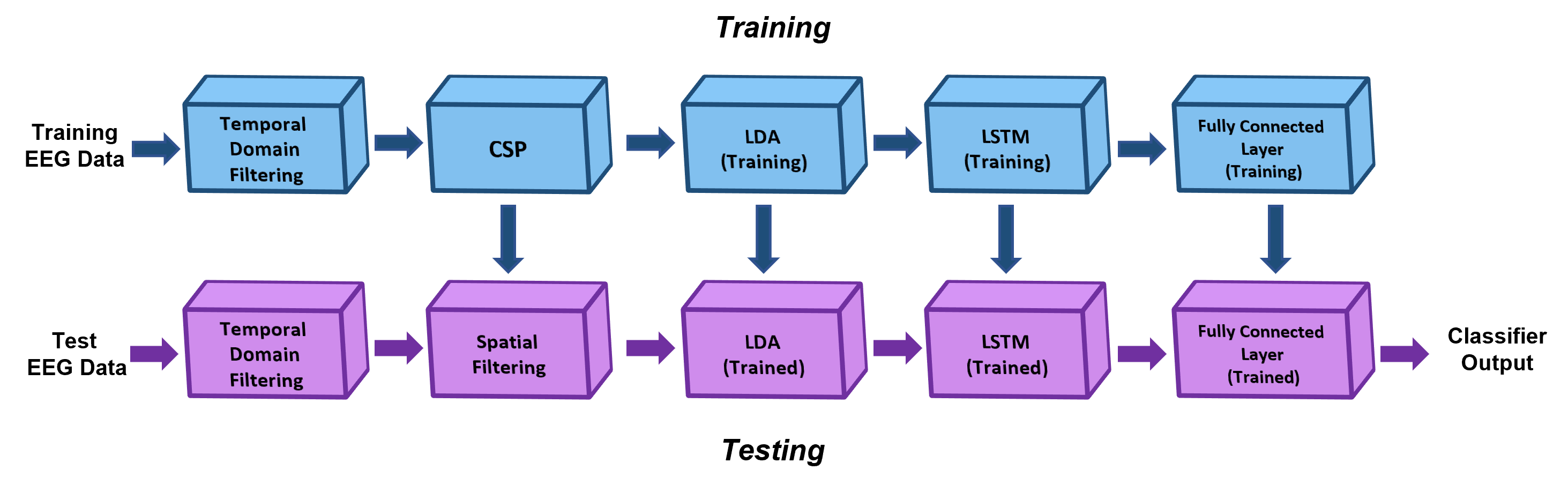}
    \caption{CSP-LDA-LSTM architecture which employs common spatial pattern (CSP) and long short-term memory (LSTM) for classifying meditative-state EEG epochs from the non-meditative ones. Linear discriminant analysis (LDA) is used for dimensionality reduction.}
    \label{fig:arch}
\end{figure}
\subsection{CSP-LDA-LSTM Architecture}
The CSP-LDA-LSTM architecture is shown in Fig. \ref{fig:arch}. Since there was no performance improvement with regularization in CSP-LDA architecture, we chose the classical CSP for this architecture. Ten spatial filter pairs were chosen since there was no improvement in accuracy with more filter pairs. To create adequate samples to train the LSTM classifier, the epoch length was reduced from 256 to 64 samples. The logarithm of variance of each filtered signal is utilized as a feature, which results in a 20-dimensional feature vector. LDA reduces the dimension of these vectors to one. The resulting values are presented as sequence data to a single-layer LSTM with 200 hidden units, and to obviate overfitting, the number of training epochs is limited to 20. Adam optimizer is employed, and the learning rate used is 0.001. The difference between the architectures of the classifiers employed in this study and the OPTICAL predictor used in \cite{kumar2019brain} is that LSTM is used as the classifier in the former, while it is employed in a regression setting in \cite{kumar2019brain}, where support vector machine is the classifier.

\begin{figure}
    \centering
    \includegraphics[width = 8cm]{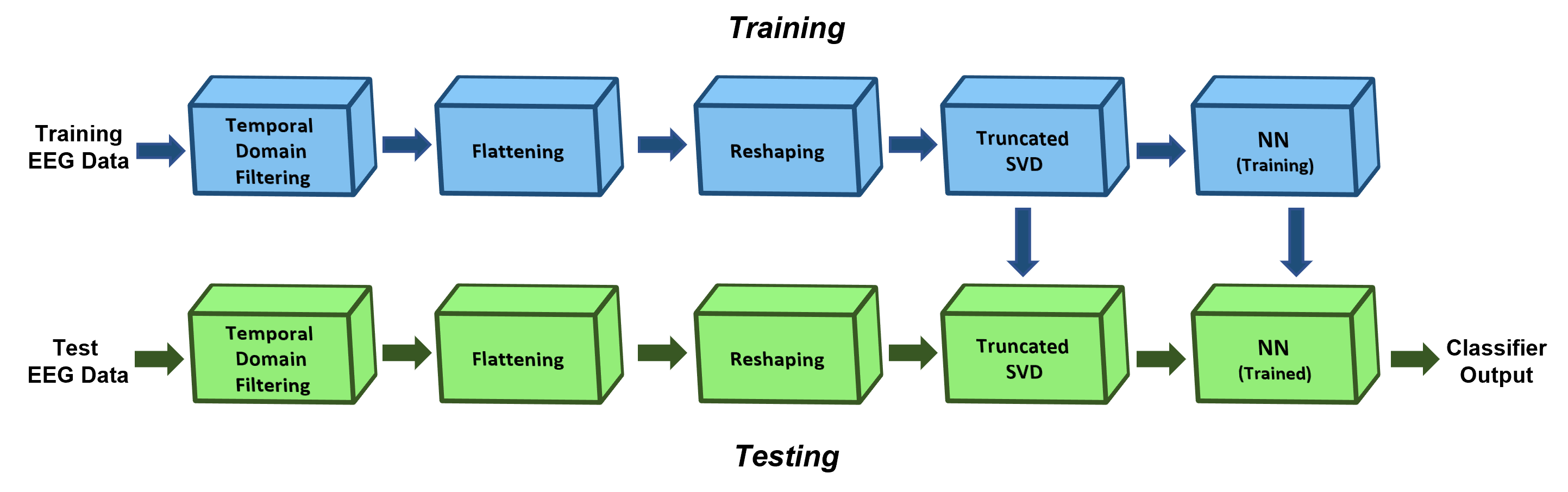}
    \caption{SVD-NN architecture employing singular value decomposition (SVD) and shallow neural network (NN) for classifying meditative state EEG epochs from non-meditative ones.}
    \label{fig:arch2}
\end{figure}
\subsection{SVD-NN Architecture}
Figure \ref{fig:arch2} gives the block diagram of the SVD-NN architecture. Singular value decomposition is an elegant mathematical tool for extracting crucial information (dimensionality reduction) and signal analysis primarily in the fields of image \cite{kakarala2001signal,selvan2007svd,mateen2019fundus,sethares2014eigentextures} and speech processing \cite{xue2019robust,winursito2018feature,uhl2001experiments,panachakel2017high}. SVD is also used in studies involving EEG for classification of motor imagery \cite{baali2015transform,mesbah2015motor}, emotion  \cite{veena2020human}, epileptic seizure detection  \cite{zhang2018generalized,parvez2013classification}, and assessment of drivers' cognitive distraction \cite{almahasneh2014eeg}.

SVD factorizes a real matrix $\mathbf{X}$ of rank $r$ as,
\begin{eqnarray}
    \mathbf{X} = \mathbf{U}\mathbf{\Sigma}\mathbf{V^T} = \sum_{i=1}^{r}\sigma_i\mathbf{u}_i\mathbf{v}_i^T
\label{svdEq}
\end{eqnarray}
where the orthogonal matrices $\mathbf{U}$ and $\mathbf{V}$ contain the eigenvectors of $\mathbf{XX^T}$ and $\mathbf{X^TX}$, respectively and the rectangular diagonal matrix $\mathbf{\Sigma}$ contains the $r$ non-zero singular values ($\sigma_i$) of $\mathbf{X}$ as the diagonal entries. $\mathbf{u}_i$ and $\mathbf{v}_i$, respectively, denote the $i^{th}$ columns of $\mathbf{U}$ and $\mathbf{V}$. 

EEG reconstructed from the truncated SVD, which is a low-rank approximation of the original signal, is used as the input to NN. The rationale behind this is the assumption that the original EEG signal is ``noisy'' and the low-rank approximation is its denoised version, provided the noise is distributed uniformly across all the orthogonal directions given by the tensors $\mathbf{u}_i\mathbf{v}_i^T$. If $\mathbf{X}$ is corrupted with an additive noise distributed across the directions of the tensors $\mathbf{u}_i\mathbf{v}_i^T$, the signal-to-noise ratio (SNR) of each component of the expansion in Eq. \ref{svdEq} decreases with decreasing values of $\sigma_i$. That is, if $SNR_i$ denotes the SNR in the $i^{th}$ component in Eq. \ref{svdEq}, then $SNR_k \ge SNR_j$ if $\sigma_k \ge \sigma_j$. If we replace $r$ in Eq. \ref{svdEq} with $k$ where $k < r$, then instead of $\mathbf{X}$, we get a rank-$k$ approximation of $\mathbf{X}$, say $\mathbf{\Tilde{X}}$, as a denoised version of $\mathbf{X}$. It is reasonable to assume that the actual electrophysiological signal corresponding to meditation lies in a subspace whose dimension is less than that of the recorded EEG.

$\mathbf{X}$ is constructed by the following procedure:

\begin{itemize}
 \item Step 1: Two-dimensional EEG signals of a given epoch (of length 128 samples) are flattened column-wise into a single-column vector.
 \item Step 2: Column vectors formed from different epochs of the same class (meditation or resting state) are stacked together to form a matrix $\mathbf{X}$.
 \item Step 3: Matrix $\mathbf{X}$ is reshaped to have 384 columns. This value has been arrived at after experimenting with different multiples of 128.
\end{itemize}

The matrix $\mathbf{X}$ is decomposed using SVD and reconstructed using a limited number of larger singular values (SVs), chosen based on validation set performance. In intra-subject classification, 10-fold cross-validation is performed, with 90\% of one subject's data used for training and 10\% for testing in each fold. A randomly selected 10\% of the training data from the first fold serves as the validation set to determine the optimal number of SVs, which is then used for all folds.

For inter-subject classification, leave-one-subject-out cross-validation is used, with data from 53 subjects for training and one for testing in each round. A validation set, comprising 10\% of the training data from the first round, is used to select the optimal number of SVs for all rounds. The reconstructed vectors from these SVs are input to a shallow neural network with two hidden layers: 64 or 32 neurons in the first layer (depending on input size) and 8 neurons in the second. ReLU activation is used in the hidden layers, sigmoid for classification, with Adam optimizer (learning rate: 0.0004) and binary cross-entropy loss.
\section{Results and Discussion}
\label{res}
The efficiency of all the proposed approaches for classifying the resting from the meditative-state EEG epochs is tested under both inter-subject and intra-subject settings. In intra-subject settings, the training and testing data arise from the same subject and ten-fold cross-validation is resorted to. In the inter-subject setting, training is performed using the data from $N-1$ meditators and testing, using the remaining subject in every cross-validation round. Thus the number of crossvalidation steps is the same as the number of meditators studied, and the test data comes from a subject not used while training the classifier. Achieving satisfactory inter-subject classification performance is possible only if the classifier can learn subject-independent features.

\begin{figure}
    \centering
    \includegraphics[width = 8cm]{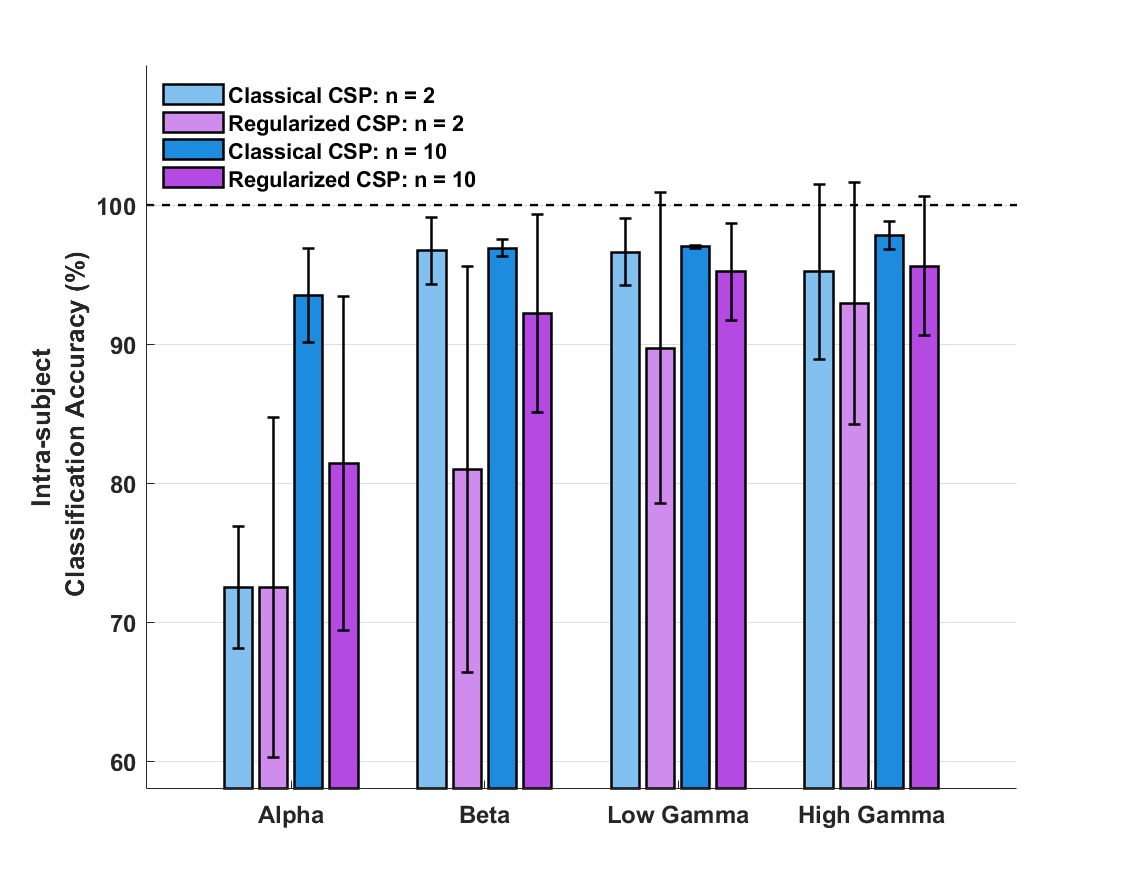}
    \caption{Comparison of the intra-subject classification performance of the CSP-LDA architecture for each EEG frequency band. The graph shows the results for classical CSP and TR-CSP (regularized). The y-axis gives the mean accuracy values (in \%) obtained during 10-fold cross-validation. $n$ is the number of filter pairs used. The error bars show the standard deviation. The total number of meditators: 54.}
    \label{fig:intracomb}
\end{figure}

\begin{figure}
    \centering
    \includegraphics[width = 8cm]{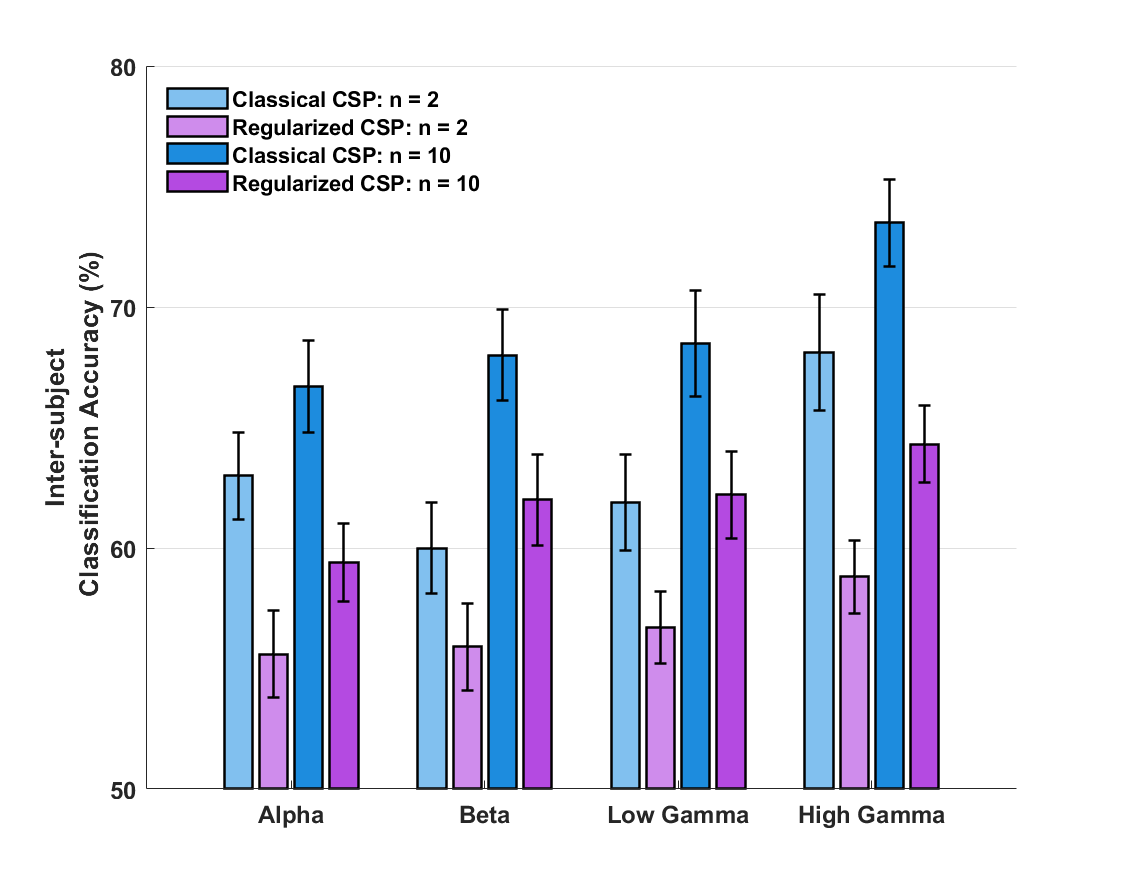}
    \caption{Comparison of the inter-subject classification performance of the CSP-LDA architecture for each EEG frequency band. The graph compares the results of classical CSP with those of TR-CSP (regularized). The Y-axis gives the mean accuracy (in \%) obtained by leave-one-out cross-validation. $n$ is the number of filter pairs used. The error bars show the standard deviation. The total number of meditators: 54.}
    \label{fig:intercomb}
\end{figure}

Figure \ref{fig:intracomb} gives the intra-subject classification results with each EEG frequency band, when CSP-LDA architecture is used. The classical CSP performs better than regularized CSP. Also, we see an improvement in the accuracy with an increase in the number of filter pairs. 

Figure \ref{fig:intercomb} shows the inter-subject classification results employing CSP-LDA architecture. Similar to intra-subject classification, we see a clear advantage for classical CSP. The effect of an increase in the number of filter pairs is not as pronounced in the inter-subject classification setting as in the intra-subject setting. 

The results obtained using the CSP-LDA-LSTM architecture in both intra-subject and inter-subject classification settings are given in Table \ref{clltab}. Similar to the CSP-LDA architecture, we see a reduction in the performance in the inter-subject setting compared to the intra-subject setting. We also see a performance improvement when high-frequency bands are used. This observation is in line with the results of the CSP-LDA architecture. The highest intra-subject accuracy obtained using the CSP-LDA-LSTM architecture is $98.2\pm0.5$\% for the high-gamma band and for inter-subject classification, it is $94.1 \pm 0.4$\%.

\begin{table}
\centering
\caption{Performance of CSP-LDA-LSTM architecture in classifying meditative from the resting state in the intra-subject and inter-subject classification settings. The total number of meditators: 54.}
\label{clltab}
\begin{tabular}{@{} cccc @{}}
\toprule
\textbf{Sl. No.} & \textbf{EEG Frequency} & \textbf{Intra-subject} & \textbf{Inter-subject} \\
 & \textbf{Band} & \textbf{Accuracy (\%)} & \textbf{Accuracy (\%)} \\
\midrule
1                & Alpha                       & $95.6 \pm 0.6$                       & $79.1 \pm 0.7$                       \\
2                & Beta                        & $97.1 \pm 0.2$                       & $86.5 \pm 0.6$                       \\
3                & Low Gamma                   & $97.9 \pm 0.1$                       & $90.7 \pm 0.5$                       \\
4                & High Gamma                  & $\mathbf{98.2 \pm 0.5}$              & $\mathbf{94.1 \pm 0.4}$              \\
\bottomrule
\end{tabular}
\end{table}

The performance of the SVD-NN architecture in intra-subject and inter-subject settings are given in Tables \ref{optintra} and \ref{optinter}, respectively. Unlike the CSP-LDA and CSP-LDA-LSTM architectures, the best performance with SVD-NN is observed for the beta band and not for the high gamma band. The highest accuracy is $97.7 \pm 1.4$\% for intra-subject classification and $96.4\pm 0.6$\% for inter-subject setting.

\begin{table}[h!]
\centering
\caption{Intra-subject accuracy values of SVD-NN architecture for classifying meditative from the rest state. The evaluation strategy is 10-fold cross-validation. The number of singular values, which gives the highest mean accuracy on the validation set, is also listed. The total number of meditators: 54.}
\label{optintra}
\begin{adjustbox}{max width=10cm}
\begin{tabular}{@{}cccc@{}}
\toprule
\textbf{Sl. No.} & \textbf{EEG Frequency} & \textbf{Optimal Number} & \textbf{Accuracy} \\
& \textbf{Band} & \textbf{of Singular Values} & \textbf{(\%)} \\
\midrule
1 & Alpha & 132 & $93.1 \pm 1.9$ \\
2 & Beta & 252 & $\mathbf{97.7 \pm 1.4}$ \\
3 & Low Gamma & 119 & $52.1 \pm 3.4$ \\
4 & High Gamma & 122 & $59.8 \pm 15.2$ \\
\bottomrule
\end{tabular}
\end{adjustbox}
\end{table}

\begin{table}[h!]
\centering
\caption{Inter-subject performance of SVD-NN architecture in classifying meditative from the resting state. The evaluation strategy is leave-one-out cross-validation. The number of singular values resulting in the highest mean accuracy on the validation set is also listed. The total number of meditators: 54.}
\label{optinter}
\begin{tabular}{@{}cccc@{}}
\toprule
\textbf{Sl. No.} & \textbf{EEG Frequency} & \textbf{Optimal Number} & \textbf{Accuracy (\%)} \\
& \textbf{Band} & \textbf{of Singular Values} & \\
\midrule
1 & Alpha & 127 & $91.8 \pm 1.7$ \\ 
2 & Beta & 247 & $\mathbf{96.4 \pm 0.6}$ \\ 
3 & Low Gamma & 124 & $51.4 \pm 11.3$ \\ 
4 & High Gamma & 129 & $39.2 \pm 12.7$ \\ 
\bottomrule
\end{tabular}
\end{table}

 Table \ref{tabcomp2} compare the performance of CSP-LDA, CSP-LDA-LSTM, and SVD-NN architectures in the intra-subject and inter-subject classification settings, for different EEG frequency bands. Table \ref{tabcomp} compares the performance of the proposed methods with those of the techniques in the literature. However, it must be noted that each of the three techniques we have compared with from the literature has dealt with a different type of meditation, and not the Rajyoga meditation studied by us. This is because there is no study reported in the literature on Rajyoga meditation for the problem addressed in this study. The highest accuracy obtained in the intra-subject classification setting is $98.2 \pm 0.5$\% when the CSP-LDA-LSTM architecture is used with high gamma band. This is marginally higher than the state-of-the-art performance of 97.9\% obtained by Lin and Li \cite{lin2017using} on Chan Buddhist meditation practitioners. Though our results are comparable with those of Lin and Li \cite{lin2017using}, we have achieved these results using epochs of length 2 s whereas they had used epochs of length 10 s. 

The highest inter-subject classification accuracy obtained is $96.4 \pm 0.6$\% using the SVD-NN architecture with the beta band. Since no study in the literature has reported subject-independent performance, we are unable to compare our results with any other result in the literature.

\begin{table*}[h!]
\centering
\caption{Performance comparison of all the proposed architectures for different EEG frequency bands. For the CSP-LDA architecture, the accuracies reported are for classical CSP with 10 pairs of spatial filters. The highest accuracy in each band and each classification setting is given in boldface.} 
\label{tabcomp2}
\begin{adjustbox}{width = 16cm}
\begin{tabular}{cc|cccccccc}
\toprule
                                       &                       & \multicolumn{8}{c}{\textbf{Performance for different EEG Frequency Bands (in \%)}}                                            \\ 
\midrule
                                       &                       & \multicolumn{2}{c}{\textbf{Alpha Band}}                                                                                                                                                                         & \multicolumn{2}{c}{\textbf{Beta Band}}                                                                                                                                                                          & \multicolumn{2}{c}{\textbf{Low Gamma Band}}                                                                                                                                                                     & \multicolumn{2}{c}{\textbf{High Gamma Band}}                                                                                                                                               \\ 
\cmidrule{3-10}
\textbf{Sl. No.} & \textbf{Architecture} & \textbf{\begin{tabular}[c]{@{}c@{}}Intra-Subject \\ Accuracy\end{tabular}} & \textbf{\begin{tabular}[c]{@{}c@{}}Inter-Subject \\ Accuracy\end{tabular}} & \textbf{\begin{tabular}[c]{@{}c@{}}Intra-Subject \\ Accuracy\end{tabular}} & \textbf{\begin{tabular}[c]{@{}c@{}}Inter-Subject \\ Accuracy\end{tabular}} & \textbf{\begin{tabular}[c]{@{}c@{}}Intra-Subject \\ Accuracy\end{tabular}} & \textbf{\begin{tabular}[c]{@{}c@{}}Inter-Subject \\ Accuracy\end{tabular}} & \textbf{\begin{tabular}[c]{@{}c@{}}Intra-Subject \\ Accuracy\end{tabular}} & \textbf{\begin{tabular}[c]{@{}c@{}}Inter-Subject \\ Accuracy\end{tabular}} \\ 
\midrule
1                & CSP-LDA               & $93.5 \pm 3.4$                                                                     & $66.7 \pm 1.9$                                                                    & $96.9 \pm 0.6$                                                                     & $68.0 \pm 1.9$                                                                    & $97.0 \pm 0.1$                                                                     & $68.5 \pm 2.2$                                                                    & $97.8 \pm 1.0$                                                                     & $73.5 \pm 1.8$                                                                    \\ 
2                & CSP-LDA-LSTM          & $\mathbf{95.6 \pm 0.6}$                                                            & $79.1 \pm 0.7$                                                                   & $97.1 \pm 0.2$                                                                     & $86.5 \pm 0.6$                                                                    & $\mathbf{97.9 \pm 0.1}$                                                            & $\mathbf{90.7 \pm 0.5}$                                                           & $\mathbf{98.2 \pm 0.5}$                                                            & $\mathbf{94.1 \pm 0.4}$                                                           \\ 
3                & SVD-NN                & $93.1 \pm 1.9$                                                                     & $\mathbf{91.8 \pm 1.7}$                                                          & $\mathbf{97.7 \pm 1.4}$                                                            & $\mathbf{96.4 \pm 0.6}$                                                           & $52.1 \pm 3.4$                                                                     & $51.4\pm 11.3$                                                                    & $59.8 \pm 15.2$                                                                    & $39.2 \pm 12.7$                                                                   \\ 
\bottomrule
\end{tabular}
\end{adjustbox}
\end{table*}

\begin{table*}
\centering
\caption{Performance comparison of our methods with the techniques in the literature involving other meditation practices. Lin and Li \cite{lin2017using} have used ten-second epochs, whereas we used two-second epochs. DWT: discrete wavelet transform; SVM: support vector machine; CSP: common spatial pattern; LDA: linear discriminant analysis; LSTM: long short-term memory; SVD: singular value decomposition; NN: shallow neural network; NA: No. of subjects not given in the paper.} 
\label{tabcomp}
\begin{tabular}{@{}cccccccc@{}}
\toprule
\textbf{Sl. No.} & \multicolumn{1}{c}{\textbf{Technique}} & \multicolumn{1}{c}{\textbf{Med. type}} &
\multicolumn{1}{c}{\textbf{Intra/inter}} &
\multicolumn{1}{c}{\textbf{No. of subj.}} &
\multicolumn{1}{c}{\textbf{Features}} & \multicolumn{1}{c}{\textbf{Classifier}} & \multicolumn{1}{c}{\textbf{Max. Accuracy}} \\ \midrule
           1      &  Tee at al. \cite{tee2020classification}                                  &  Theta healing                                               &  Intra-subject & 20 
 & DWT                                   &       Logistic regression                               &                             96.9\%                    \\
           2      & Ahani et al. \cite{ahani2014quantitative} &     Mindfulness                                           &       Intra-subject                               &  34 
 &  Stockwell transform  &       SVM                                  &    78.0\%                                            \\
            3     & Lin and Li \cite{lin2017using}                                     &    Chan                                               &  Intra-subject &  NA  &             Approximate entropy                         &     Bagged tree                                     &     97.9\%                                                                                                                        \\
                       4      &  CSP-LDA                                  & Rajyoga                                                  &  Intra-subject & 54 &  CSP-high gamma                                      &      LDA                                    &   97.8\%                                             \\ 
                       5      &  CSP-LDA-LSTM                                  & Rajyoga                                                  &  Intra-subject &  54  & CSP-high gamma                                      &      LSTM                                    &   \textbf{98.2\%}                                             \\ 
                       6      &  SVD-NN                                   & Rajyoga                                                  &  Intra-subject & 54 & SVD-beta                                      &      NN                                    &   97.7\%                                             \\  \hline
                       7      &    CSP-LDA                                      & Rajyoga                                                 &  Inter-subject & 54  & CSP-high gamma                                      &      LDA                                    &   73.5\%                                             \\
                       8      &  CSP-LDA-LSTM                                   & Rajyoga                                                  &  Inter-subject & 54  &  CSP-high gamma                                     &      LSTM                                    &   94.1\%                                             \\
                       9      &  SVD-NN                                    & Rajyoga                                                  &  Inter-subject & 54  & SVD-beta                                      &      NN                                    &   \textbf{96.4\%}                                            \\
\bottomrule
\end{tabular}
\end{table*}
\subsection{Effect of Different Frequency Bands}
For CSP-LDA and CSP-LDA-LSTM architectures, higher performance is achieved with higher frequency bands. Conversely, the SVD-NN architecture performs better with lower frequency bands, with the optimal number of singular values varying significantly across bands. These values, chosen based on validation set accuracy, are listed in Tables \ref{optintra} and \ref{optinter}. For the alpha band, the best singular values are 132 (intra-subject) and 127 (inter-subject), while for the beta band, they are approximately double at 252 and 247, respectively. This suggests that the beta band captures more distinguishing information between meditative and resting states, whereas the alpha band may include common neural activities that reduce separability.

Gamma band classification accuracies approach chance levels, indicating limited discriminative power. The contradictory results between CSP-based architectures and SVD-NN may stem from differences in feature extraction and classifiers. CSP relies on selected channels optimizing an objective function, while SVD-NN utilizes all channels. Additionally, as noted by Pandey et al. \cite{pandey2021brain}, the discriminability of EEG features can depend on the channels used, further influencing the results. Overall, all frequency bands provide useful information, with the optimal band depending on the architecture and features used.
\subsection{Effect of TR Regularisation}
Although Lotte et.al  \cite{Lotte2010} have reported that the performance of the system improves with regularisation, in our experiments, regularisation has not improved the accuracy. Consistently across all the EEG frequency bands and classification settings, classical CSP without regularisation has performed as well as the regularized CSP. This might be because Tikhonov regularization penalizes higher-valued elements in the spatial vector.
\subsection{Effect of the Number of Filter Pairs Employed}
Classification accuracy initially improves for all the frequency bands as the number of filter pairs employed increases but remains constant for higher values. This is distinct from what has been reported earlier \cite{panachakel2021decoding} for the task of classifying imagined words using deep neural network. It has been reported that the accuracy increases till the number of filter pairs reaches nine but instead of plateauing, performance deteriorates with a further increase in the number of filters.
\subsection{Effectiveness for other meditative practices}
The effectiveness of our features and architectures needs to be experimentally tested to distinguish other types of meditation from the resting state of subjects. Since most types of meditations are practiced with eyes closed, networks trained on EEG data obtained from Rajyoga meditators (which is practiced with eyes open) may not directly work on data from practitioners of other meditative practices. Thus, our methods need to be separately tested for their performance on EEG data obtained from practitioners of other types of meditations.
\section{Conclusion}
Three distinct architectures are presented for classifying the Rajyoga meditation state from an eyes-open baseline. The first uses EEG power differences, the second applies sequence learning techniques, and the third projects data into a lower-dimensional subspace before classification. The CSP-LDA-LSTM architecture achieved $98.2 \pm 0.5\%$ accuracy in intra-subject experiments using the high-gamma band, outperforming prior works. The SVD-NN architecture reported $96.4 \pm 0.6\%$ accuracy in inter-subject classification with the beta band, marking a significant subject-independent result.

Results highlight that the best EEG frequency band depends on the classification problem and method used. CSP-based architectures perform better with the high-gamma band, suggesting notable gamma-band power differences between meditation and rest. SVD-NN performs best with the beta band, likely due to its ability to exclude neural activities from the alpha band in the occipital lobe during eyes-open meditation. Both architectures demonstrate practical accuracy for meditative state classification.

{
\bibliographystyle{IEEEtran}
\bibliography{MILE.bib}
}

\end{document}